\documentclass[10pt,twocolumn,letterpaper]{article}

\usepackage{cvpr}
\usepackage{times}
\usepackage{graphicx}
\usepackage{amsmath}
\usepackage{amssymb}

\usepackage[tight,footnotesize,sf,SF]{subfigure}

\usepackage{makecell}
\usepackage{booktabs}
\usepackage{floatrow}
 \usepackage{enumerate}

\DeclareFloatFont{tiny}{\footnotesize}%
\usepackage[labelfont=bf,labelsep=period,font=small]{caption}

\usepackage{mdwlist}
\usepackage{nopageno}

\usepackage{capt-of,etoolbox} %

\newcommand{\includetikz}[1]{%
    \includegraphics{tikzFigs/{#1}.pdf}
}

\usepackage[pagebackref=true,breaklinks=true,colorlinks,bookmarks=false]{hyperref}

\cvprfinalcopy %

\newcommand{\R}{\mathbb{R}}

\newcommand{\bigO}{\mathcal{O}}
\newcommand{\vecEll}{\boldsymbol{\tau}}
\newcommand{\vecTau}{\boldsymbol{\tau}}

\newcommand{\med}{\operatorname{median}}
\newcommand{\id}{\operatorname{Id}}

\newcommand{\cE}{\mathcal{E}}

\newcommand{\IN}{\mathbb{N}}
\newcommand{\IR}{\mathbb{R}}
\newcommand{\IS}{\mathbb{S}}

\newcommand{\dx}{{\mathrm{d}x}}

\DeclareMathOperator{\SE}{SE}
\DeclareMathOperator{\SO}{SO}

\newcommand{\abs}[1]{{\left|#1\right|}}

\newtheorem{lemma}{Lemma}
\newtheorem{proposition}{Proposition}
\newtheorem{definition}{Definition}

\def\cvprPaperID{0346} %

\ifcvprfinal\fi

\begin{document}

\title{A Combinatorial Solution to Non-Rigid 3D Shape-to-Image Matching}

\author{Florian Bernard\textsuperscript{1,2,3}$\qquad$
Frank R. Schmidt\textsuperscript{3}$\qquad$
Johan Thunberg\textsuperscript{1}$\qquad$
Daniel Cremers\textsuperscript{3}
\\
$\newline$
\\
\textsuperscript{1}Luxembourg Centre for Systems Biomedicine, University of Luxembourg, Luxembourg\\
\textsuperscript{2}Centre Hospitalier de Luxembourg, Luxembourg \\
\textsuperscript{3}Technical University of Munich (TUM), Germany
}

\makeatletter
\let\@oldmaketitle\@maketitle%
\renewcommand{\@maketitle}{\@oldmaketitle%
  \myfigure{}\bigskip}%
\makeatother

\newcommand\myfigure{%
  \centering \vspace{-6mm}
  \includegraphics[scale=0.1825]{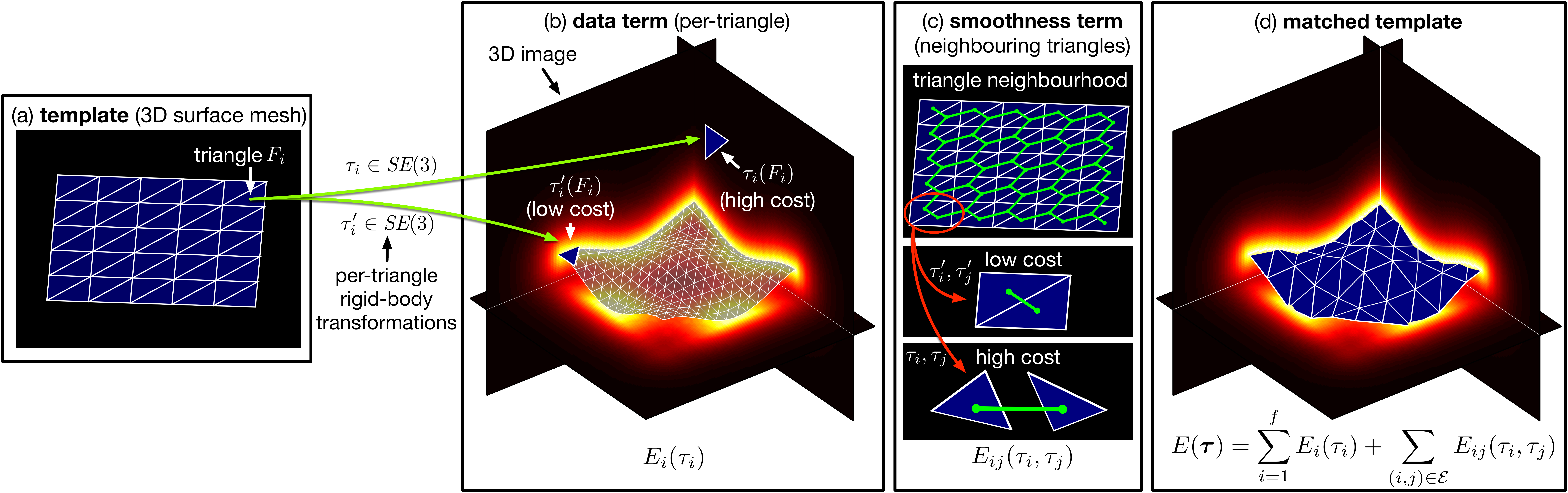}
  \vspace{-6mm}
    \bf\captionof{figure}{\rm~3D Shape-to-Image Matching. %
        (a)~Given a shape as a triangular mesh, we associate to each
        triangle $F_i$ its rigid transformation $\tau_i\in\SE(3)$
        such that the sum of data and smoothness terms is minimised.
        (b)~The data term $E_i(\tau_i)$ measures how well the transformed triangle
        $\tau_i(F_i)$ fits into the volumetric image. (c)~The smoothness
        term $E_{ij}(\tau_i,\tau_j)$ penalises the discrepancy between
        transformed triangles $\tau_i(F_i)$ and $\tau_j(F_j)$. (d)~Optimising $E$ provides
        us with a shape-to-image matching. %
    }
    \label{teaser}
  } 

\maketitle

\begin{abstract}%
We propose a combinatorial solution for the problem of non-rigidly matching a 3D shape to 3D image data. To this end, we model the shape as a triangular mesh and allow each triangle of this mesh to be rigidly transformed to achieve a suitable matching to the image. By penalising the distance and the relative rotation between neighbouring triangles our matching compromises between image and shape information. In this paper, we resolve two major challenges: Firstly, we address the resulting large and NP-hard combinatorial problem with a suitable graph-theoretic approach.  Secondly, we propose an efficient discretisation of the unbounded 6-dimensional Lie group $\SE(3)$. To our knowledge this is the first combinatorial formulation for non-rigid 3D shape-to-image matching.  In contrast to existing local (gradient descent) optimisation methods, we obtain solutions that do not require a good initialisation and that are within a bound of the optimal solution.  We evaluate the proposed method on the two problems of non-rigid 3D shape-to-shape and non-rigid 3D shape-to-image registration and demonstrate that it provides promising results.%
\end{abstract}

\section{Introduction}
Matching a shape template to an image is a well studied problem in
computer vision and image analysis. It gives rise to a wide range of
applications, including image segmentation and object detection. An
early approach for the detection of lines and parametrised curves in
images is the voting-based Hough transform \cite{Duda:1972uj}, which
was later generalised to the detection of arbitrary shapes
\cite{ballard1981generalizing}.

Whilst the Hough transform considers rigid shapes, the utilisation of
shape information in image segmentation tasks has also been addressed
in the non-rigid case, including methods based on active shape models
\cite{Cootes:1992uw}, level sets \cite{Cremers:2006wv}, convex shape
spaces \cite{Cremers:2008tl}, multiphase graph cuts \cite{Vu:2008jv},
or statistical shape models \cite{Heimann:2009kv,Zhang:2011gu}.  For
the reconstruction of the shape of an object from a single 2D image,
shape-from-template approaches aim to match a given 3D template to the
image via a 3D-to-2D projection
\cite{Salzmann:2009er,Malti:2015vu,Parashar:2015va}. Several authors
have considered combinatorial formulations of the non-rigid shape-to-image matching problem for certain classes of shapes. For the case of
matching contours
\cite{Coughlan:2000dw,Schoenemann:2010ug}, or 2D
chordal graph polygons \cite{Felzenszwalb:2005ep}, the resulting
optimisation problems can be solved globally.  However, a
generalisation of these methods to 3D shapes is non-trivial and we are
not aware of previous work addressing this issue.  The purpose of this
work is to fill this gap by presenting a combinatorial formulation for
the non-rigid matching of a 3D shape template to a 3D image. For that,
we model the shape as a triangular mesh and allow each triangle
$F_i$ of this mesh to be independently transformed via a rigid
transformation $\tau_i\in\SE(3)$.  
Using a discretisation of the unbounded 6-dimensional Lie group
$\SE(3)$, we formulate the matching task as a manifold-valued
multi-labelling problem that can be cast as minimising the energy
\begin{align}
  E(\vecTau)=\sum_{i=1}^n
  E_i(\tau_i)+\sum_{(i,j)\in\cE}E_{ij}(\tau_i,\tau_j) .
  \label{energyFcn}
\end{align}
Here, the data term $E_i(\tau_i)$ takes the image information into
account, while the smoothness term $E_{ij}(\tau_i,\tau_j)$ measures
the dissimilarity between the observed shape and the modelled shape
prior. By penalising the distance and the relative rotation of
neighbouring triangles our matching compromises between image and
shape information. %
In general, minimising functions of the form in~\eqref{energyFcn} is
NP-hard \cite{Kolmogorov:2004he}.%

\subsection{Related Work}
  To the best of our knowledge the present
  paper is the first one that considers a combinatorial formulation of
  the non-rigid 3D shape to 3D image matching problem.  In the
  following we will summarise methodologies that are most relevant to
  our work.

{\textbf{Continuous Optimisation:} In many scenarios it is
  natural to assume that image or shape deformations are spatially
  continuous and smooth. Frequently, such problems are formulated in
  terms of optimisation problems over the space of diffeomorphisms
  \cite{Dupuis:1998ur,Michor:2004vw,Beg:2005wy,Miller:2006ft}.
  Commonly, gradient descent-like methods are used to obtain (local)
  optima of the (typically non-convex) problems.
  However, a major shortcoming of these methods is that a good initial
  estimate is crucial and in general there are no bounds on the
  optimality of the solution. To deal with the non-convexity of a 2D
  shape-to-image matching problem that is formulated in terms of
  optimal transport, the authors in \cite{Schmitzer:2015uo} propose to
  use a branch and bound scheme.
}

\textbf{Shortest Paths and Dynamic Programming:} In contrast to the
continuous local optimisation methods, many vision problems can be
formulated in a discrete manner such that they are amenable to
solutions based on graph algorithms and dynamic programming (DP)
\cite{Felzenszwalb:2011fg}.  Since curves are intrinsically
one-dimensional, various curve matching formulations can also be
reduced to finding a shortest-path in a particular graph.  Moreover,
based on a recursive formulation using easier-to-solve subproblems,
matching problems with templates that have a tree structure can frequently be tackled by DP.
For a deformable matching of an open contour
to a 2D image, a global solution based on DP has been proposed in
\cite{Coughlan:2000dw}. Also based on DP, in
\cite{Felzenszwalb:2005ep} the authors present a method for solving
the problem of deformably matching a 2D polygon to a 2D image for
chordal graph polygons.  In
\cite{Schoenemann:2010ug}, the authors propose a
globally optimal approach for matching a closed contour to a 2D image
based on cycles in a product graph of the contour and the image. A
related formulation that is also based on a product graph has recently
been introduced in \cite{Lahner:2016vp} for deformable contour to 3D
shape matching.

\textbf{Graph-cuts:} It is well known (see e.g.~\cite{Boykov:2003uy}) that any cut of a graph can be
  interpreted as finding a closed manifold of co-dimension 1 in the
  ambient space (e.g., closed curve in 2D, closed surface in 3D,
  etc.).  
  One such example
is the reconstruction of a 3D shape 
  from a set of sparse 3D points, where the latter is
represented on a discrete 3D grid \cite{Lempitsky:2007hl}.

\textbf{Labelling Problems:} Labelling problems are ubiquitous in
computer vision and appear both in continuous and discrete settings
\cite{Zach:2014fb}. The popular Markov Random Field (MRF) framework
offers a Bayesian treatment thereof \cite{Li:2012wj}. Also, linear
programming relaxations of MRFs have been studied
\cite{Werner:2007ws}. The continuous approaches to multi-labelling
include various convex
relaxations~\cite{Pock:2008vf,Lellmann:2011vf,Strekalovskiy:2011fz,Goldluecke:2013gh},
multi-labelling problems with total variation regularisation of
functions with values on manifolds~\cite{Lellmann:2013ka}, as well as
sublabel-accurate convex
relaxations~\cite{moellenhoff-laude-cvpr16,laude16eccv}. Among the
discrete multi-labelling methods are the previously-mentioned
graph-cuts, which can be used to find global solutions for certain
binary labelling problems, including problems with submodular pairwise
costs \cite{kolmogorov2007minimizing}. %
For a sub-class of multi-labelling problems a global solution can also
be found \cite{kolmogorov2007minimizing}. This sub-class includes
pairwise costs that are convex in terms of totally ordered labels
\cite{Ishikawa:2003wl}. In addition, efficient algorithms for finding
local optima of general multi-labelling problems have been proposed
\cite{Boykov:2001ug,Komodakis:2007wr}, which even have theoretical
optimality guarantees. A more detailed description of the energy
functions that can be optimised using graph-cuts is given in
\cite{Kolmogorov:2004he,Kolmogorov:2005vm,kolmogorov2007minimizing}.

\subsection{Main Contributions}
The main contribution of this paper is to present for the first time a
combinatorial formulation of the non-rigid 3D shape to 3D image
matching problem. Whilst our problem is a natural extension to the
afore-mentioned ``dimension one'' matching approaches
\cite{Coughlan:2000dw,Felzenszwalb:2005ep,Schoenemann:2010ug,Lahner:2016vp},
a generalisation to (intrinsic) \emph{dimension two} problems is more
intricate.
Our main contributions are:
\begin{itemize}
\item By using a surface mesh transformation model that makes use of
  \emph{per-triangle} rigid transformations, we formulate the 3D shape to
  3D image matching problem in terms of a manifold-valued
  multi-labelling problem.
\item We introduce a pairwise term that defines a metric on the label
  space $\SE(3)$, which itself is a high-dimensional Lie group.  With
  that, our energy function is amenable to be minimised by the
  $\alpha$-expansion algorithm \cite{Boykov:2001ug}, which has been shown
  to work well in practice, is efficient even for very large label
  spaces, and has theoretical optimality guarantees.
\item In contrast to continuous optimisation methods that use
  gradient descent-like algorithms, our combinatorial method does
  not require a good initialisation.
\item In order to deal with the computationally challenging
  discretisation of $\SE(3)$, we propose to use a coarse-to-fine
  discretisation of the Lie group.
\end{itemize}

\section{Non-Rigid 3D Shape-to-Image Matching}\label{sec:methods}
In this section we first specify our objective, followed by a
description of the data term and smoothness term. After introducing
the combinatorial problem, we describe the
discretisation of the label space and we discuss the algorithmic
solution of the problem.

\subsection{Objective Function}\label{objFcn}

In the following, we assume that a 3D shape $S\subset\IR^3$ is given
as a triangular mesh. This means we have $n\in\IN$ triangles
$F_1,\ldots,F_n\subset\IR^3$ such that 
\begin{align}
  S = \bigcup_{i=1}^n F_i.
\end{align}

  We use the set $\mathcal{E} \subset \{1,\ldots,n\}^2$
  to define the neighbourhood
  between pairs of (different) triangles.
  We assume that for all
  $(i,j) \in \mathcal{E}$ the neighbouring triangles $F_i$ and $F_j$
  are non-disjoint and that the intersection $F_i\cap F_j$ results
  either in a common edge or a common vertex. Also, w.l.o.g. we assume
  that for each $(i,j) \in \mathcal{E}$ it holds that $i<j$, \ie
  $(i,j) \in \mathcal{E} \Rightarrow (j,i) \notin \mathcal{E}$.

  Our objective is it now to match the 3D shape $S$ onto a volumetric
  image $I\colon\Omega\to\IR^c$, where $\Omega\subset\IR^3$ denotes the compact
    image domain and $c\in\IN$ describes the amount of image
  channels.  While we are interested in a non-rigid shape-to-image
  matching, we like to favour matchings that are as-rigid-as-possible,
  similar to the approach in \cite{Sorkine:2007wt} that applies
  (locally regularised) rigid transformations to \emph{each vertex}.
  However, in our case this is done by applying to \emph{each
    triangle} $F_i$ a rigid transformation
\begin{align}
  \tau_i=(\tilde{\tau}_i,\vec{\tau}_i)\in\SE(3)=\SO(3)\ltimes\IR^3,
\end{align}
where $\tilde{\tau}_i\in\SO(3)\subset\IR^{3\times 3}$ represents the
rotational part and $\vec{\tau}_i\in\IR^3$ represents the
translational part of $\tau_i$.
The task of finding the best matching $\vecTau =
(\tau_1,\ldots,\tau_n)$ can be formulated as minimising the energy
\begin{align}
  E(\vecTau)=\sum_{i=1}^n E_i(\tau_i)+\sum_{(i,j) \in
    \mathcal{E}}E_{ij}(\tau_i,\tau_j) .
\end{align}
 In Section~\ref{sec:E_data}
we define the data term $E_i(\tau_i)$ that evaluates how well the
transformed triangle $\tau_i(F_i)$ fits to the image data.  In
Section~\ref{sec:E_smoothness} we define the smoothness term
$E_{ij}(\tau_i,\tau_j)$ that measures the geometric dissimilarity
between the shape model $S$ and the transformed shape%
\begin{align}
  \vecTau(S) := \bigcup_{i=1}^n \tau_i(F_i).
\end{align}
Using the proposed piecewise rigid transformation model we 
  may end up with a model
  $\vecTau(S)$ having
(small) gaps or intersections between neighbouring triangles. We will
later address this issue and present a simple yet effective way of
dealing with this irregularity.

\subsection{Data Term}\label{sec:E_data}
The data term $E_i(\tau_i)\in\IR$ measures how well the transformed
triangle $\tau_i(F_i)$ fits to the image data $I$. 
For that, we introduce the \emph{score image} $J:\Omega \rightarrow
[0,1]$ that is derived from the image $I$ (e.g. a gradient magnitude
image, or more advanced predictors based on neural networks).  For a
triangle $F \subset \Omega$, we define 
\begin{align}
  J[F]:=\int_F J(x)\dx. \label{Jfcn}
\end{align}
 With that, the value $J[F]$
indicates how well the triangle $F$ fits to the image data, where a high value of the score image indicates a
good fit. 
The data term is then given by
\begin{align}
  E_i(\tau_i) = -J[\tau_i(F_i)]. 
\end{align}
In the discrete setting, the data term $E_i$ is computed by a weighted
sum of function values $-J(x)$ over the triangle. The weights take the rasterisation of
the deformed triangle $\tau_i(F_i)$ in the image into account.

\subsection{Smoothness Term}\label{sec:E_smoothness}
The pairwise term $E_{ij}(\tau_i,\tau_j)\in\IR_0^+$ penalises the
disagreement between neighbouring triangles $F_i$ and $F_j$ after they have
been transformed by $\tau_i$ and $\tau_j$, respectively.

For defining the pairwise term we first introduce suitable distances.
The expression 
\begin{align}
  d_{\SO(3)}(\tilde{\tau}_i, \tilde{\tau}_j) &= 
  \sqrt{\frac{1}{2}}\left\|\log(\tilde{\tau}_i^T \tilde{\tau}_j)\right\|_F \label{distso3log}
\end{align}
is the geodesic distance between the rotations $\tilde{\tau}_i$ and
$\tilde{\tau}_j$ on $\SO(3)$ \cite{Huynh:2009bw}, with matrix logarithm $\log(\cdot)$.
For $q_i$ and $q_j$ being the
  quaternion representations of $\tilde{\tau}_i$ and $\tilde{\tau}_j$,
  one can efficiently compute the distance as $d_{\SO(3)} = 2\cos^{-1}(\abs{\langle q_i,
    q_j \rangle})$, where $\langle \cdot, \cdot \rangle$ is the quaternion
  inner-product \cite{Huynh:2009bw}.

For defining a distance between neighbouring triangles, we make use of the concept of group actions. To be more specific, we define 
\begin{align}
  d_{\SE(3),X}({\tau}_i, {\tau}_j) =  \max_{x \in X}\left\|  \tau_i(x) {-} \tau_j(x) \right\|_2, \label{dStretch}
\end{align}
where the group $\SE(3)$ acts on the non-empty compact set $X \subseteq \R^3$. In our case we
use $X = F_i \cap F_j$ such that $d_{\SE(3),F_i \cap F_j}({\tau}_i,
{\tau}_j)$ can be seen as distance between the deformed triangles
$\tau_i(F_i)$ and $\tau_j(F_j)$. In this case the maximum in
$d_{\SE(3),F_i \cap F_j}({\tau}_i, {\tau}_j)$ is achieved at
the common vertices of $F_i$ and $F_j$, which is attractive from a
computational point of view.

Using the introduced distances, we define our pairwise term as a
weighted sum thereof, \ie
\begin{align}
  &E_{ij}(\tau_i,\tau_j) =\label{pairwiseTerm}  \\
   &\qquad\quad\lambda_B d_{\SO(3)}(\tilde{\tau}_i,
  \tilde{\tau}_j)  
   + \lambda_S d_{\SE(3),F_i \cap F_j}({\tau}_i,{\tau}_j) .\nonumber
\end{align}
The purpose of the bending term, weighted by $\lambda_B > 0$, is to
ensure that the rotations of neighbouring triangles are similar.  The
stretching term, weighted by $\lambda_S > 0$, ensures that
neighbouring triangles stay close together.

\subsection{Combinatorial Formulation}\label{combForm}
A matching $\vecTau$ of shape $S$ to the image $I$ is given by a
solution of the optimisation problem
\begin{align}
  \min_{\substack{\vecTau \in \SE(3)^n}} & \quad
  E(\vecTau).\label{contProb}
\end{align}
Due to the non-convexity of the feasible set $\SE(3)^n$, it follows
that Problem~\eqref{contProb} is non-convex. This non-convexity makes
it difficult to solve the problem directly over the unbounded
continuous space $\SE(3)^n$. Our approach is to optimise
instead over a discretisation of the search space. With that, we
obtain a multi-labelling problem, for which efficient and effective
algorithms are
available.  %
  For the discretisation of $\SE(3)$ we make use of the
fact that it is a product space of $\SO(3)$ and $\R^3$. Thus, we
define $\mathcal{L} \subset \SO(3) \ltimes \R^3 = \SE(3)$ to be the
(finite) manifold-valued label space that contains $\ell = \vert
\mathcal{L} \vert$ elements of $\SE(3)$.  
\textbf{Translations:}
The Lie group $\SE(3)$ is non-compact due to the translational part being
encoded by $\R^3$. However, since the image domain $\Omega$ is
compact, the image size provides natural bounds for a discretisation
of the translations.
Let $n_x, n_y, n_z$ be the number of voxels of the image $I$ and
$\ell_x, \ell_y,\ell_z$ be the number of labels for the $x$, $y$, $z$
translations. For convenience, and w.l.o.g., we assume that our
template is defined relative to the centre of the image domain, \ie
 the template's centre-of-gravity coincides with the centre
of the image. Moreover, w.l.o.g., we assume that we are looking for a
matching such that a substantial part of the (transformed) template
lies inside the image\footnote{If this is not the case, one can increase
  the image size accordingly.}.  Let us define $\mathcal{Z}_m(n) = \{-\frac{n}{2},
\ldots, 0, \ldots, \frac{n}{2}\}$ to be the set containing $m$
evenly-spaced elements with centre $0$, where $m$ is an odd positive
integer. The \emph{diameter} $n$ defines the difference between the
largest and the smallest elements.
  A
discretisation of the translations is given by the set
$\vec{\mathcal{L}} = \mathcal{Z}_{\ell_x}(n_x) \times
\mathcal{Z}_{\ell_y}(n_y) \times \mathcal{Z}_{\ell_z}(n_z)$ with
$\vert \vec{\mathcal{L}} \vert = \ell_x {\cdot} \ell_y {\cdot}
\ell_z$.  
\textbf{Rotations:} Various works that are related to the
discretisation of $\SO(3)$ have previously been presented. These
include sampling strategies for rigid-body path planning
\cite{Kuffner:2004tl}, an approximation of the neighbourhood in
$\SO(3)$ based on vector distances \cite{Li:2007ey}, or an analysis of
various %
metrics for 3D rotations \cite{Huynh:2009bw}.
Our discretisation of $\SO(3)$ is based on the \emph{Hopf fibration},
which describes $\SO(3)$ in terms of the circle $\IS^1$ and the
2-sphere $\IS^2$. The intuition of this approach is to transfer a
discretisation of $\IS^1$ and $\IS^2$ to the space of rotations.
We refer the interested reader to
\cite{Yershova:2010ij} for a detailed description.
Let $\tilde{\mathcal{L}}$ denote the so-obtained set of a uniform sampling of $\SO(3)$ containing $\tilde{\ell} = \vert \tilde{\mathcal{L}} \vert$ elements. 
By optimising $E$ over the label space $\mathcal{L}^n$, we now obtain the combinatorial optimisation problem as
\begin{align}
  \min_{\substack{\vecTau \in \mathcal{L}^n}} & \quad
  E(\vecTau).\label{discProb}
\end{align}

\section{Algorithm}
In order to solve Problem~\eqref{discProb}, we use $\alpha$-expansion~\cite{Boykov:2001ug,Kolmogorov:2004he,Boykov:2004hg},
which greedily updates only one label at a time. We note that
there are also potential alternatives to $\alpha$-expansion (e.g.~for
non-metric pairwise
terms~\cite{Komodakis:2007wr,Strekalovskiy:2011fz}, or fusion
moves~\cite{Lempitsky:2010wf}).  Whilst $\alpha$-expansion has the requirement that the pairwise
term is a
metric, %
it is appealing both from a practical and a theoretical point of view.
To be more specific, it is efficient, robust with respect to
initialisation, the obtained local optimum is guaranteed to lie within
a factor of the global optimum, and an efficient implementation that
supports the online computation of the smoothness term is
available~\cite{Boykov:2001ug,Kolmogorov:2004he,Boykov:2004hg}, which
is crucial for the size of problems that we are solving.
We now show that our pairwise term is a metric and thus
$\alpha$-expansion is applicable.
\begin{lemma} \label{propMetric} %
Let $X \subseteq \R^3$ be a non-empty compact set and $\tau_i, \tau_j, \tau_k \in \SE(3)$. The stretching term $$d_{\SE(3),X}({\tau}_i, {\tau}_j) =  \max_{x \in X}\left\|  \tau_i(x) {-} \tau_j(x) \right\|_2,$$ is a pseudometric, \ie it satisfies
\begin{enumerate}[(i)]
  \item $d_{\SE(3),X}(\tau_i, \tau_j) \geq 0$,~ $d_{\SE(3),X}(\tau_i, \tau_i) = 0$,
  \item symmetry: $d_{\SE(3),X}({\tau}_i, {\tau}_j) = d_{\SE(3),X}({\tau}_j, {\tau}_i)$, and
  \item the triangle inequality:\\
  ${d_{\SE(3),X}({\tau}_i, {\tau}_k) \leq d_{\SE(3),X}({\tau}_i, {\tau}_j) {+} d_{\SE(3),X}({\tau}_j, {\tau}_k).}$
\end{enumerate}
\end{lemma}

\noindent\textit{Proof:} %
\noindent (i) and (ii) follow directly from the definition. The triangle inequality holds, since%
\begin{align*}
  & d_{\SE(3),X}({\tau}_i, {\tau}_k) = \max_{x \in X}\left\|  \tau_i(x) {-} \tau_k(x) \right\|_2\\
   & ~~=  \max_{x \in X}\left\|  \tau_i(x) - \tau_k(x)  + \tau_j(x) - \tau_j(x)\right\|_2 \\
     & ~~\leq \max_{x \in X} \left( \left\|  \tau_i(x) - \tau_j(x)\right\|_2 + \left\| \tau_j(x) {-} \tau_k(x) \right\|_2 \right)\\
   & ~~\leq  \max_{x \in X} \left\|  \tau_i(x) - \tau_j(x)\right\|_2 + \max_{x \in X} \left\| \tau_j(x) {-} \tau_k(x) \right\|_2 \\
   &~~=d_{\SE(3),X}({\tau}_i, {\tau}_j) + d_{\SE(3),X}({\tau}_j, {\tau}_k). \qquad\qquad~ \qquad\blacksquare
\end{align*}

\begin{proposition}
  For $\lambda_S, \lambda_B > 0$ and $(i,j) \in \mathcal{E}$, the pairwise term $E_{ij}(\cdot,\cdot)$ defined in~\eqref{pairwiseTerm} is a metric. %
\end{proposition}

\noindent\textit{Proof:} Due to the assumption in
Section~\ref{objFcn}, for $(i,j) \in \mathcal{E}$ it follows that $X
:= F_i \cap F_j \neq \emptyset$ is compact. Thus, $d_{\SE(3),X}(\cdot,\cdot)$ is
a pseudometric (Lemma~\ref{propMetric}). Whilst
$d_{\SO(3)}(\cdot,\cdot)$ is known to be a metric on $\SO(3)$, it is
only a pseudometric on $\SE(3)$. Since $E_{ij}(\cdot,\cdot)$ is a
positive linear combination of two pseudometrics,
$E_{ij}(\cdot,\cdot)$ is also a pseudometric. To show that
$E_{ij}(\cdot,\cdot)$ is a metric, we show that $E_{ij}(\tau_i,\tau_j)
{=} 0$ implies $\tau_i{=}\tau_j$. For $E_{ij}(\tau_i,\tau_j) {=} 0$,
it holds that $d_{\SO(3)}(\tilde{\tau}_i,\tilde{\tau}_j) {=} 0$, which
implies $\tilde{\tau}_i {=}\tilde{\tau}_j$. Moreover, with
$E_{ij}(\tau_i,\tau_j) {=} 0$ and $\tilde{\tau}_i {=}\tilde{\tau}_j$,
it holds that $d_{\SE(3),X}(\tau_i,\tau_j) {=} \max_{x \in X}\left\|
  (\tilde{\tau}_i(x) {+}\vec{\tau}_i) {-} (\tilde{\tau}_j(x) {+}
  \vec{\tau}_j ) \right\|_2 = \left\| \vec{\tau}_i {-} \vec{\tau}_j
\right\|_2 {=} 0$, which implies $\vec{\tau}_i {=}\vec{\tau}_j$. Hence
$\tau_i {=}\tau_j$. \hfill $\blacksquare$

\subsection{Coarse-to-Fine Processing}
In practice, for a reasonably large number of labels $\ell = \tilde{\ell} {\cdot} \ell_x {\cdot} \ell_y {\cdot}
\ell_z$, a direct solution of
Problem~\eqref{discProb} is intractable.
In order to cope with this issue we propose to use a
coarse-to-fine strategy that (approximately) solves Problem~\eqref{discProb} at
different levels $s$ of the label space. Let $s{=}0$ 
  denote the coarsest (initial) level and
$s{=}s_{\max} \geq 0$ the finest (final) level. Once a solution
$\vecEll^{(s)}$ has been obtained at level $s$, for running the algorithm at
level $s{+}1$ the labelling is initialised with $\vecEll^{(s)}$ and
the label space is updated accordingly. %
For computational efficiency, in the coarse-to-fine approach each
triangle has its own feasible label space. 
Let $\mathcal{L}^{(s)}_i$
denote this feasible label space for the $i$-th triangle at level $s$. Initially, on the base level $s{=}0$, the label spaces are the same for each triangle,
\ie
$\mathcal{L}^{(0)}_i = \mathcal{L}^{(0)}$. The general idea for obtaining $\mathcal{L}^{(s+1)}_i$ is to consider a (uniform) discretisation of
 the neighbourhood of the transformation $\tau_i^{(s)}$ at level $s$, where the radius of the neighbourhood decreases across the levels.
Let us introduce a
neighbourhood on $\SO(3)$:

\begin{definition} ($\epsilon$-ball on $\SO(3)$)\\
  The ball on $\SO(3)$ with radius $\epsilon$ and centre $\tilde{\tau}
  \in \SO(3)$ is defined as
  $\mathcal{B}_{\epsilon}^{\SO(3)}(\tilde{\tau}) = \{ \tau \in \SO(3)
  : d_{\SO(3)}(\tau,\tilde{\tau}) < \epsilon \}$.
\end{definition}

Next, we describe the coarse-to-fine structure of the label space
based on its product space nature. Since each label can be
written as 
\begin{align}
  \tau_i^{(s)} = (\tilde{\tau}_i^{\,(s)}, \vec{\tau}_i^{~(s)}) \in
  \tilde{\mathcal{L}}^{\,(s)}_i \times \vec{\mathcal{L}}^{~(s)}_i
  \subset \SO(3) \ltimes \R^3,
\end{align}
we can consider the translations and rotations independently. Note that $\tilde{\mathcal{L}}^{\,(s)}_i \times \vec{\mathcal{L}}^{~(s)}_i \subset \SE(3)$ is not necessarily a group anymore. By enforcing that $\tau_i^{(s)} \in \mathcal{L}^{(s+1)}_i$, the solution $\vecTau^{(s)}$ at level $s$ is also contained in the new label space.
Thus, the energy cannot increase from level $s$ to $s{+}1$.

\textbf{Translations:} We define $\vec{\mathcal{L}}^{(0)} := \vec{\mathcal{L}}$ (cf.~Section~\ref{combForm}).
For obtaining the set of translations at level $s{+}1$ for triangle
$i$, the new translation grid at level $s{+}1$ is centred at
$\vec{\tau}_i^{~(s)}$. Moreover, the diameter from level $s$ is
reduced by a factor of two, leading to
\begin{align}
   &\vec{\mathcal{L} }_i^{~(s+1)} := \\
   &\qquad\vec{\tau}_i^{~(s)} {+} \left(\mathcal{Z}_{\ell_x}(\frac{n_x^{(s)}}{2}) {\times} \mathcal{Z}_{\ell_y}(\frac{n_y^{(s)}}{2}) {\times} \mathcal{Z}_{\ell_z}(\frac{n_z^{(s)}}{2})\right),  \nonumber
\end{align}
where the vector-set addition is element-wise. 
Initially, $n_x^{(0)} =n_x, n_y^{(0)}=n_y$ and $n_z^{(0)}=n_z$.%

\textbf{Rotations:} 
Let $\tilde{\mathcal{L}}^{[r]}$ denote a discretisation
  of (the entire) $\SO(3)$ at resolution $r$ containing
$\tilde{\ell}^{[r]}$ elements, where $\tilde{\ell}^{[r]}$ increases with increasing $r$.
$\tilde{\mathcal{L}}^{[r]}$ should
not be confused with $\tilde{\mathcal{L}}^{(s)}_i$, which is the (rotation) label
space of the $i$-th triangle at level $s$ that is to be defined below.
Following the construction in \cite{Yershova:2010ij}, we obtain $5$ resolutions of the $\SO(3)$ discretisation 
$\tilde{\mathcal{L}}^{[r]}$ for $r=0,\ldots,4$. After including the identity in $\tilde{\mathcal{L}}^{[r]}$, the number of elements ranges from $\tilde{\ell}^{[0]}=577$ to  $\tilde{\ell}^{[4]} \approx 2{\cdot} 10^6$.
Let us define a set that contains a fixed number of $p$ elements from $\tilde{\mathcal{L}}^{[r]}$ that are ``closest'' to the identity, \ie
\begin{align}
  \tilde{\mathcal{L}}^{[r]}_{p} := \tilde{\mathcal{L}}^{[r]} \cap \mathcal{B}_{\epsilon}^{\SO(3)}(\id),
\end{align}
where for each resolution $r$ the smallest $\epsilon$ that fulfils $\vert \tilde{\mathcal{L}}^{[r]}_{p} \vert \geq p$ is used as radius.

Now, we define $\tilde{\mathcal{L} }^{(0)} := \tilde{\mathcal{L}}^{[0]}$, and
\begin{align}
  \tilde{\mathcal{L} }_i^{~(s+1)} := \{\tau \tilde{\tau}_i^{\,(s)} : \tau \in \tilde{\mathcal{L}}^{[s{+}1]}_{p} \subseteq \SO(3)\},
\end{align}
which is the set of compositions of $\tilde{\tau}_i^{\,(s)}$ with all rotations in $\tilde{\mathcal{L}}^{[s{+}1]}_{p}$. We use $p = \tilde{\ell}^{[0]}=577$ for all levels $s$. For the predefined $\SO(3)$ griddings at resolutions $r=0,\ldots,4$ there always existed an $\epsilon$ such that the above inequality is tight, \ie $\vert \tilde{\mathcal{L}}^{[r]}_{p} \vert = \tilde{\ell}^{[0]} = 577$.

 By including the identity in $\tilde{\mathcal{L}}^{[s{+}1]}_{p}$, we make sure that $\tilde{\tau}_i^{~(s)} \in \tilde{\mathcal{L} }_i^{~(s+1)}$. Since $0 \in \mathcal{Z}_{\cdot}(\cdot)$, it follows that $\vec{\tau}_i^{~(s)} \in \vec{\mathcal{L} }_i^{~(s+1)}$. Thus, we have $\tau_i^{~(s)} \in \mathcal{L}_i^{~(s+1)}$, ensuring 
 that the energy cannot increase when moving from level $s$ to $s{+}1$.

\subsection{Practical Considerations}
In this section we describe some aspects for the application of the proposed method in practice. 

\subsubsection{Mesh Connectivity}
After applying an individual rigid-body transformation to each
triangle of the template mesh, in general the resulting mesh may have
gaps or intersections between neighbouring triangles. However, due to the
introduced regulariser, these gaps or intersections can be expected to
be rather small. For recovering the original mesh topology, we replace each
subset of vertices having the same position in the mesh template by their
centre-of-gravity after the transformation.

\subsubsection{Memory Requirements}
For running our algorithm we pre-compute the data term, requiring
memory of $\bigO(n {\cdot}\ell)$ (the online computation has
constant memory requirements but leads to a significantly increased
runtime).
Since pre-computing the
full pairwise term requires memory of $\bigO(n^2{\cdot}\ell^2)$,
we only precompute the bending term $d_{\SO(3)}$, 
requiring memory of $\bigO(\tilde{\ell}^2)$. The stretching term
$d_{\SE(3),X}$
is computed online.

\section{Results}\label{sec:results}
For the evaluation of our method we focus on demonstrating the general
applicability of our approach. Both point-set registration
\cite{Besl:1992iv,Rangarajan:1997un,Chui:2003wr,Li:2007ey,Myronenko:ve,Myronenko:2010wn,Jian:2011fh,Horaud:2011kz,Rasoulian:2012jqa,Maron:2016vv}
and the related correspondence problem for 3D shapes
\cite{VanKaick:2011uq,Windheuser:2011iu,Kovnatsky:2014ti,Chen:2015uj}
can be tackled using our method by recasting them as shape-to-image
matching problem.  Hence, in our evaluation below, in addition to 3D
image segmentation, we also consider the case of deformable mesh
registration.

\subsection{Deformable Mesh Registration}\label{meshReg}
In the first set of experiments we demonstrate that our method can be used to perform deformable mesh registration. To emphasise that our method is insensitive to initialisation, we compare it exemplarily with the Coherent Point Drift (CPD) algorithm \cite{Myronenko:ve,Myronenko:2010wn}, a widely-used point-cloud registration method based on Expectation Maximisation. %

\textbf{Template and Target:} For the evaluation we use a low-resolution mesh of the \emph{Stanford bunny} 
as template ($n=498$), as shown in Fig.~\ref{bunnyData} (top left). A total of $20$ deformed versions of the bunny mesh, each with a random pose, are used as registration targets. %
For that, we synthetically create deformed versions of a high-resolution bunny mesh (Fig.~\ref{bunnyData}, top right)
based on random 3D displacement vectors defined at $8$ control points on a cubic grid. These displacement vectors are then transferred to the mesh using a spline-based interpolation in order to achieve a smooth and nonlinear deformation.
One such deformed mesh is shown in Fig.~\ref{bunnyData} (bottom left). Eventually, a random pose transform is applied to the deformed shape, as shown in Fig.~\ref{bunnyData} (bottom right).
\vspace{-3mm}
\newcommand{\bunnyScale}{0.18}
\newcommand{\bunnyId}{16} %
\begin{figure}[ht!] 
\vspace{-1mm}
     \centerline{ 
        \subfigure{\includegraphics[scale=\bunnyScale]{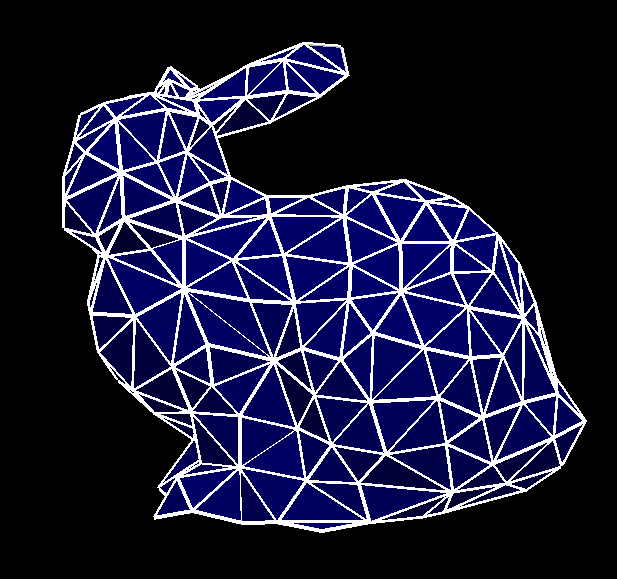} } \hfil
        \subfigure{\includegraphics[scale=\bunnyScale]{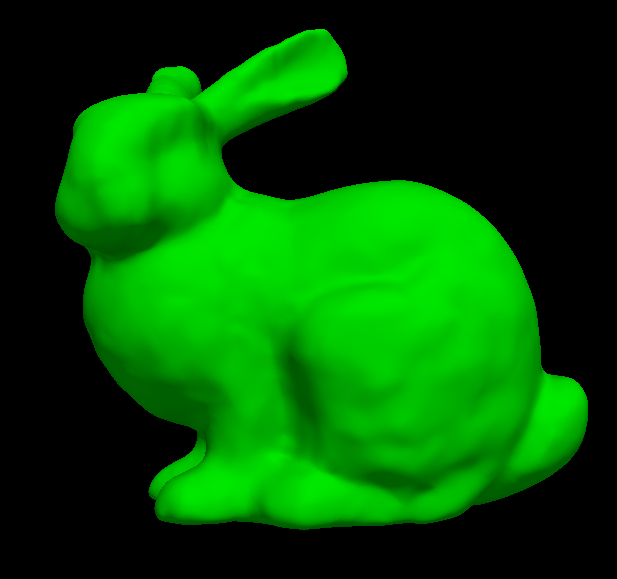} }
      }
      \vspace{-.2cm}
      \centerline{ 
        \subfigure{\includegraphics[scale=\bunnyScale]{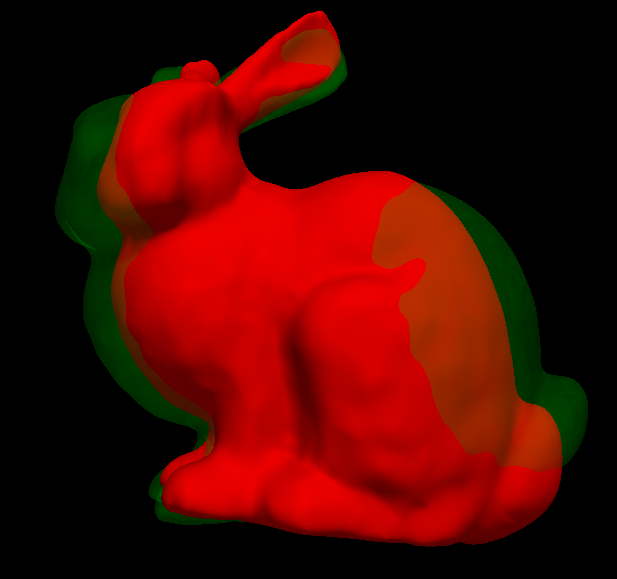} }  \hfil 
        \subfigure{\includegraphics[scale=\bunnyScale]{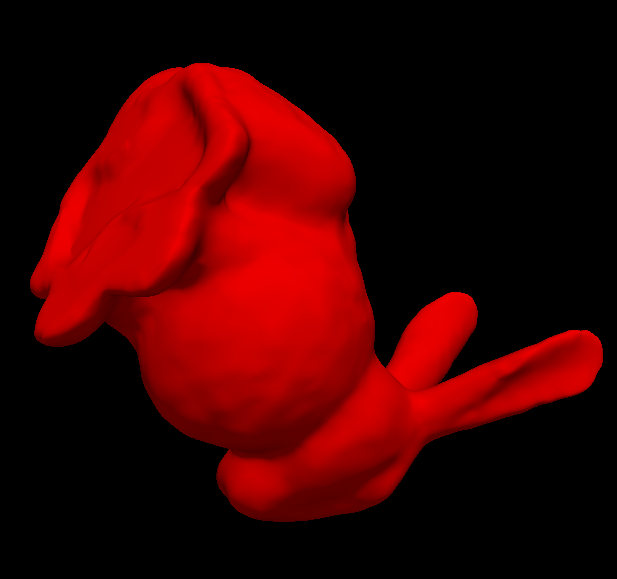} } 
     } 
     \vspace{-.45cm}
    \bf\caption{\rm~ Bunny meshes. Top left: template. Top right: high-resolution target. Bottom left: deformed target (with original target overlay). Bottom right: deformed target with random pose.} 
    \label{bunnyData}
\end{figure} 
\vspace{-4mm}
 \begin{figure}[h!] 
\vspace{-4mm}
     \centerline{
        \subfigure{\includegraphics[scale=\bunnyScale]{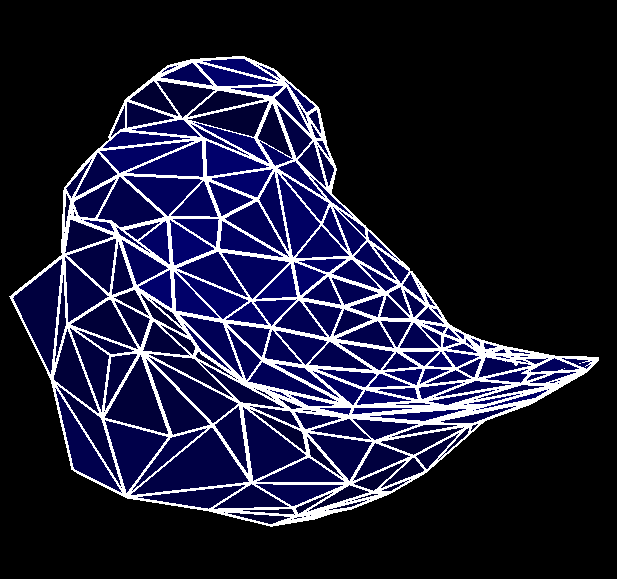} } \hfil %
         \subfigure{\includegraphics[scale=\bunnyScale]{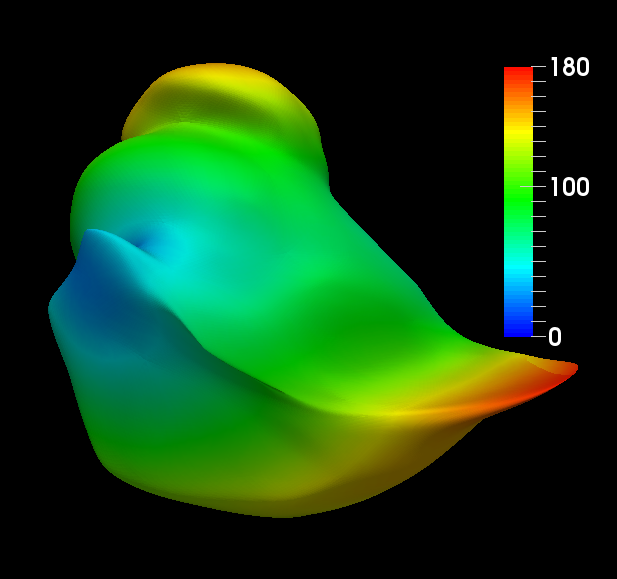} }
         }
         \vspace{-.2cm}
      \centerline{ 
           \subfigure{\includegraphics[scale=\bunnyScale]{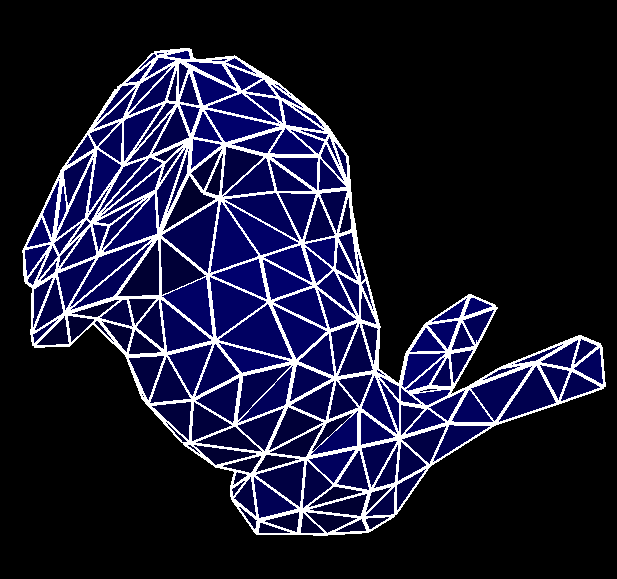} } %
            \subfigure{\includegraphics[scale=\bunnyScale]{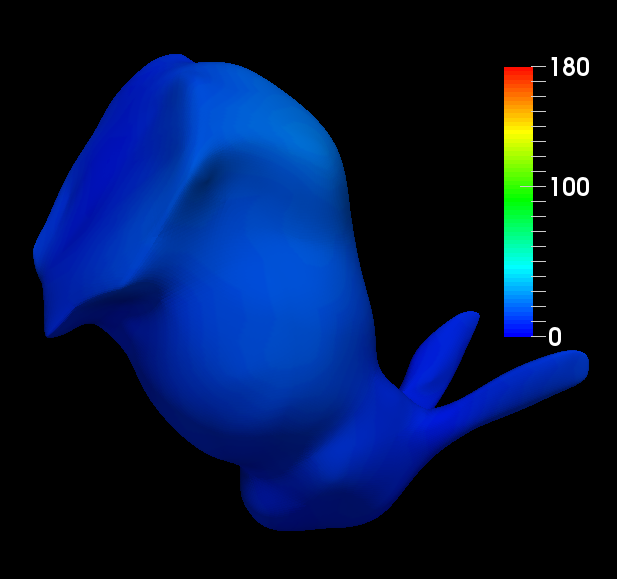} }
     } 
     \vspace{-.45cm}
    \caption{Qualitative results for registrations of the bunny template to the deformed target with random pose (cf.~Fig.~\ref{bunnyData}, bottom right). Top left: CPD result (shape destroyed). Top right: CPD error. Bottom left: our result. Bottom right: our error.} 
    \label{bunnyReg}
\end{figure} 
\vspace{-3mm}

\textbf{Score Image:} 
  In order to use our method
for mesh registration we create a score image for each target mesh and
then fit the template mesh to the score image. For $d: \Omega
\rightarrow \R^+_0$ we denote by $d(x)$ the distance of position $x$ to the boundary
of the target mesh. Now, we define the score image as $J(x) =
\exp(-\frac{d(x)}{\beta})$, where we used $\beta {=}
2$. %

\begin{figure}[h!]
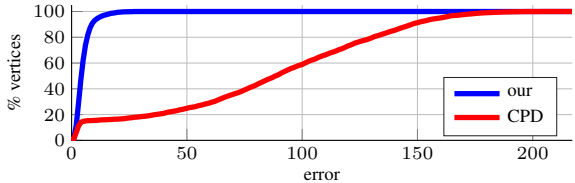

\vspace{-.1cm}
\includetikz{bunny_cumError.tikz}\hspace{0.4cm}
\caption{Percentage of vertices (vertical axis) which have an error that is smaller than or equal to the value on the horizontal axis.
}
\vspace{-.35cm}
    \label{bunnyCumError}
\end{figure}

\textbf{Parameters:} We set $\lambda_S =
54{\cdot}\frac{g}{\max(\{n_x,n_y,n_z\})} $ and $\lambda_B =
27{\cdot}\frac{g}{\pi}$, where the normalisation factor $g = \frac{n
  c_{\max}}{2 \vert \mathcal{E}\vert }$ takes the problem size into
account. The positive number $c_{\max}$ is an upper bound for the
largest possible absolute value of the data term for a single triangle
(cf.~eq.~\eqref{Jfcn}), which we compute by multiplying the largest
value of the score image by the area of the largest triangle.
The size of the label space is $\vert\mathcal{L}^{(0)}\vert = 9^3{\cdot} 577=420{,}633$, and the dimensions of the score image range from $208^3$ to $262^3$, which resulted in an average processing time of $\approx 92$ minutes per registration on a MacBook Pro (2.5GHz, 16GB).

\textbf{Results:} In the case of CPD, we first solve for a rigid registration and then for a non-rigid registration (performing a non-rigid registration directly performed worse). Since CPD is highly initialisation-sensitive it fails in $17$ out of the $20$ evaluated cases, where one representative failure case is depicted in Fig.~\ref{bunnyReg} (top row). This extreme amount of corrupt registrations emphasises the necessity of a method that is robust with respect to initialisation. In contrast, in all $20$ cases our method is able to achieve a good registration, see Fig.~\ref{bunnyReg} (bottom row) for a representative result. In Fig.~\ref{bunnyCumError} we present a quantitative evaluation.

\textbf{Discussion:} 
Whilst we do not claim to present an exhaustive evaluation of mesh
registration methods, we demonstrated the insensitivity to
initialisation of our method in a proof of concept manner.  One
advantage of our approach is that we neither have the necessity of
compatible mesh topologies, nor of compatible mesh discretisations,
since the target is represented in terms of the score image. Since the
score image is a discrete representation of the target shape
\emph{surface}, our approach amounts to a surface-based registration,
rather than a point-based registration as CPD, which is biased towards
aligning points to be as close as possible.  Moreover, the score image
offers further flexibility since additional information can be
integrated (e.g. uncertainties, mesh texture, shape features, etc.).

\subsection{Segmentation}
\begin{figure}[h!] %
     \centering{
     \includegraphics[scale=0.3]{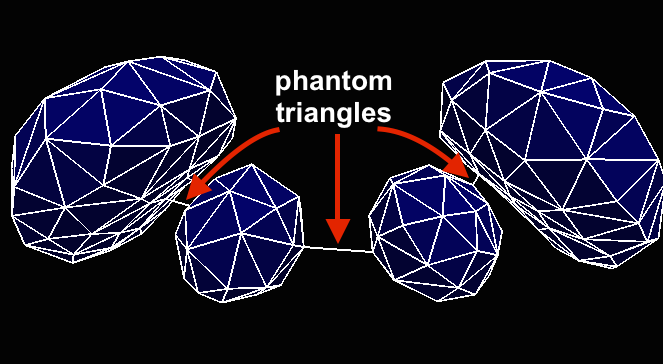}
     }
     \vspace{-3mm}
     \caption{Brain structure template.}%
     \label{bsTemplate}
\end{figure}

In the second set of experiments we apply our method to the segmentation of four brain structures (\emph{substantia nigra \& subthalamic nucleus} as single object and the \emph{nucleus ruber}, both bilaterally) in $16$ multi-modal $3$T magnetic resonance images. The delineation of the \emph{subthalamic nucleus} is known to be a challenging task, even for humans \cite{Schlaier:2011cp}. The main difficulties include weak image contrasts and the small size of the brain structures (the structures shown in Fig.~\ref{bsTemplate} are contained in a bounding box of ${\approx} 6 {\times} 3 {\times} 2.5 \text{cm}^3$, with the MRI image covering a volume of ${\approx} 20^3 \text{cm}^3$).

\textbf{Template:} For capturing the inter-relation between the brain structures, we use a multi-object template ($n{=}379$), as shown in Fig.~\ref{bsTemplate}. The template connects neighbouring brain structures by (degenerate) triangles, referred to as ``phantom triangles'', which are used only for the smoothness term and are ``free'' with respect to the data term. 
\vspace{-3mm}
\vspace{-3mm}

\textbf{Parameters:} We set $\lambda_S = 135{\cdot}\frac{g}{\max(\{n_x,n_y,n_z\})} $ and $\lambda_B = 90{\cdot}\frac{g}{\pi}$, where $g$ is defined as before. 
The size of the label space is $\vert\mathcal{L}^{(0)}\vert = {11^3{\cdot} 577}=767{,}987$, and the dimension of all score images is $364 {\times} 436 {\times} 364$, which resulted in an average processing time of $\approx 58$ minutes per fitting. %

\textbf{Score Image:} In order to perform image segmentation with our
method, we use a data term that is based on the recently proposed 3D
U-Net CNN \cite{CABR16}. For all $16$ images we train the network in a
leave-one-out manner for the prediction of volumetric segmentations.
In the centre column of Fig.~\ref{stnSeg} three examples of
so-predicted volumetric segmentations are shown. For this challenging
segmentation task the U-Net is able to identify the (rough) location
of the brain structures, but does in many cases not produce an output
that resembles the shape of the brain structures (the first two rows
in Fig.~\ref{stnSeg}). Thus, we complement the U-Net segmentations
with geometric information using our method.

For each of the four brain structures $o \in \{1,2,3,4\}$ we use an
individual score image $J_o$.  Given the binary U-Net segmentation
$I_o^{\text{unet}}: \Omega \rightarrow \{0,1\}$ for brain structure
$o$, we first extract the (predicted) boundary using morphological
operations. For $d_o:\Omega \rightarrow \R^+_0$ we denote by $d_o(x)$
the distance of position $x$ to the so-extracted boundary. Then, we
use a Gaussian kernel to define the score image as
\begin{align}
  J_o(x) = w_o \exp\left(-\frac{-d_o^2(x)}{2 \sigma_o^2}\right).
\end{align}
The weight $w_o$ and bandwidth $\sigma_o$ are used for
incorporating the confidence about the U-Net segmentation
$I_o^{\text{unet}}$ of brain structure $o$. To this end, for
\begin{align}
  \mathcal{Y}_o = \{x \in \R^3 : I_o^{\text{unet}}(x) = 1\}
\end{align}
being the one-level set of $I_o^{\text{unet}}$ that represents the point-cloud of segmented voxels, we use the average of the (per-coordinate) \emph{median absolute deviation} (MAD) as robust \emph{dispersion} measure, computed as
\begin{align}
  &\overline{\mathit{MAD}}(\mathcal{Y}_o) = \frac{1}{3}\| \mathit{MAD}(\mathcal{Y}_o)\|_1 ,\\
& \text{for~~}  \mathit{MAD}(\mathcal{Y}_o) = \med(\vert\mathcal{Y}_o - \med(\mathcal{Y}_o)\vert) ~ \in ~\R^3 . \nonumber
\end{align}
The median is understood in a per-coordinate sense, and both the
set-vector difference and the absolute value are understood
element-wise.  Now, for $\mathcal{Y}'_o$ denoting the point-cloud of
segmented voxels of structure $o$ as given by the template, the
absolute value of the average MAD difference is given by $h_o = \vert
\overline{\mathit{MAD}}(\mathcal{Y}_o) -
\overline{\mathit{MAD}}(\mathcal{Y}'_o) \vert$. With that, we define
$\sigma_o = \rho (h_o{+}1)$, scaled by $\rho{=}3$. Thus, if the
average MAD for the U-Net segmentation and the template are equal, the
bandwidth corresponds to $\rho$, whereas a larger difference in
dispersion leads to a larger bandwidth, accounting for more
uncertainty in the U-Net segmentation. Moreover, we define $w_o =
\frac{1}{h_o{+}1}$ such that an increased uncertainty in the U-Net
segmentation of brain structure $o$ leads to a decreased weight for
its data term.

\begin{figure}[ht!]
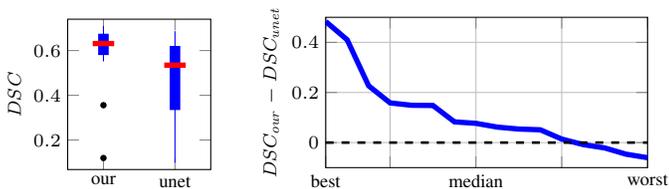

\centerline{ 
\subfigure{\includetikz{stn_DSC_boxplot.tikz}} \hspace{.35cm}
\subfigure{\includetikz{stn_DSC_difference.tikz}}
}
\vspace{-3mm}
\caption{Left: Boxplot of the DSC of our method versus the U-Net segmentation. Right: Sorted DSC differences for the $16$ cases (values above zero indicate an improvement upon U-Net).} 
    \label{stnDscDifferences}
\end{figure}

\textbf{Results:}  For the evaluation of the segmentation we use the Dice Similarity Coefficient (DSC) as volumetric overlap measure, which is defined as
  $\frac{2 \vert \mathcal{Y} \cap \mathcal{Y}' \vert}{ \vert \mathcal{Y} \vert +  \vert\mathcal{Y}' \vert}$,
for $\mathcal{Y}$ and $\mathcal{Y}'$ each being point-clouds of segmented voxels. We compute the DSC for each individual brain structure, and then report the average of the four DSC values.
Quantitative results comparing the plain U-Net segmentation and our obtained segmentations are presented in Fig.~\ref{stnDscDifferences}. The boxplot on the left reveals that in overall our method achieves higher volumetric overlaps across the $16$ cases. Moreover, the plot of sorted DSC differences on the right emphasises that applying our method improves the DSC in most cases (the values above zero), and in only a few cases it is reduced slightly.
\vspace{-3mm}
\vspace{-2mm}
In Fig.~\ref{stnSeg} qualitative results for three segmentation cases
that correspond to the \emph{best}, \emph{median} and \emph{worst}
cases in Fig.~\ref{stnDscDifferences} (right) are shown. In the first
row of Fig.~\ref{stnSeg} it can be seen that in some cases our method
is even able to achieve a reasonable segmentation based on a poor
U-Net segmentation. This was possible by putting a stronger emphasis
on the shape information relative to the data term, which also biases
the method towards the shape information. A related discussion on the
biasedness of model-to-data fitting approaches in
structure-from-motion can be found in \cite{Nurutdinova:2015kj}.
\newcommand{\stnId}{13} 
\newcommand{\stnScale}{.1} 
\vspace{-2mm}
\begin{figure}[ht!]
    \centerline{\makebox[0.3\textwidth][l]{$~~~$ground truth}\makebox[0.3\textwidth]{U-Net}\makebox[0.3\textwidth]{$~~~~~~~~~$our}}
    \vspace{-0.1cm}
     \centerline{ 
     \rotatebox{90}{$~~\quad$best}
        \subfigure{\includegraphics[scale=\stnScale]{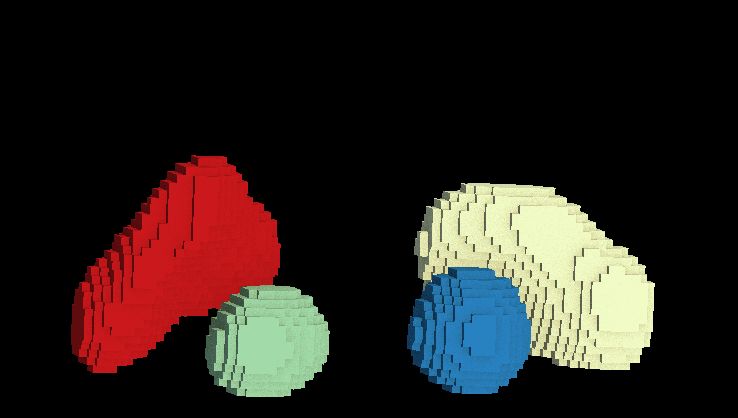} } \hfil
        \subfigure{\includegraphics[scale=\stnScale]{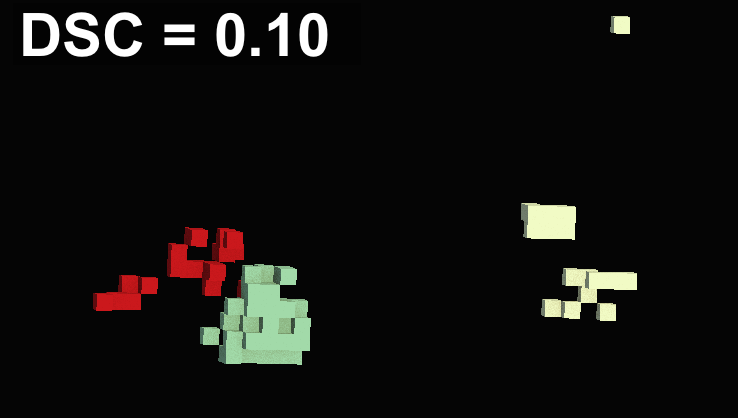} } \hfil
        \subfigure{\includegraphics[scale=\stnScale]{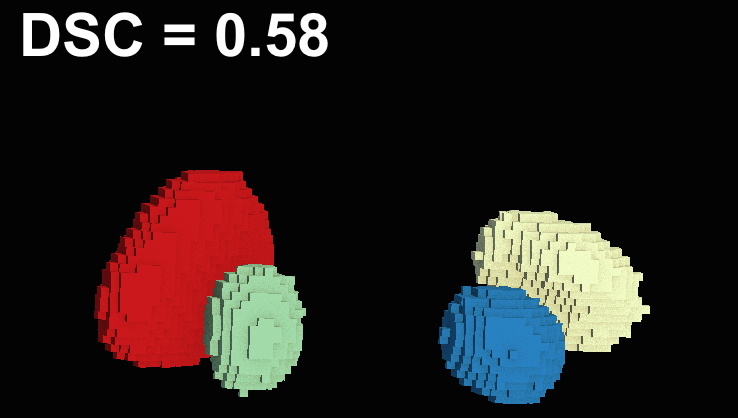}}%
     }%
     \vspace{-0.35cm}
     \centerline{ 
     \rotatebox{90}{$~~\qquad$median}
        \subfigure{\includegraphics[scale=\stnScale]{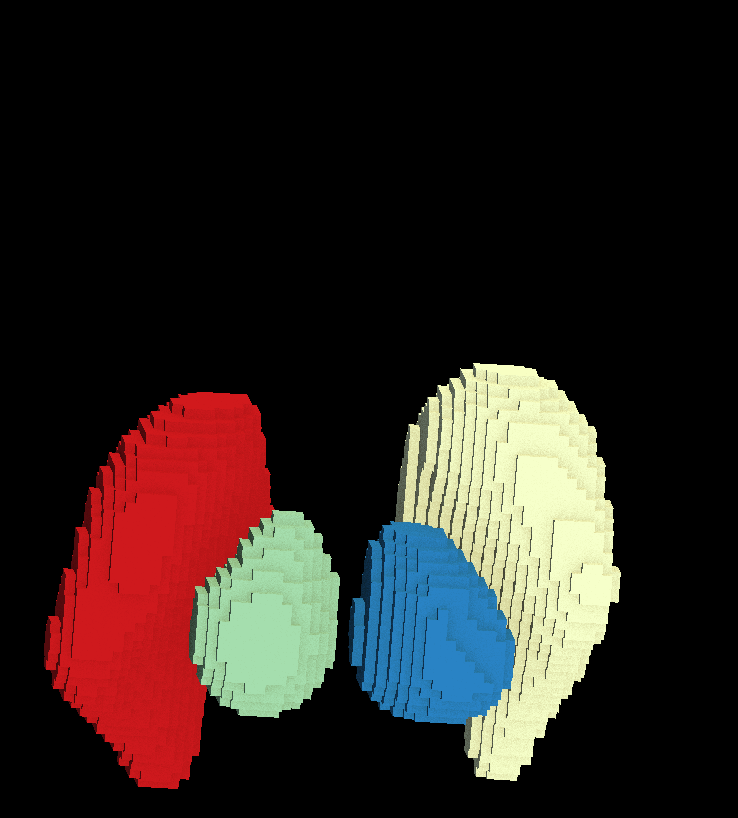} } \hfil
        \subfigure{\includegraphics[scale=\stnScale]{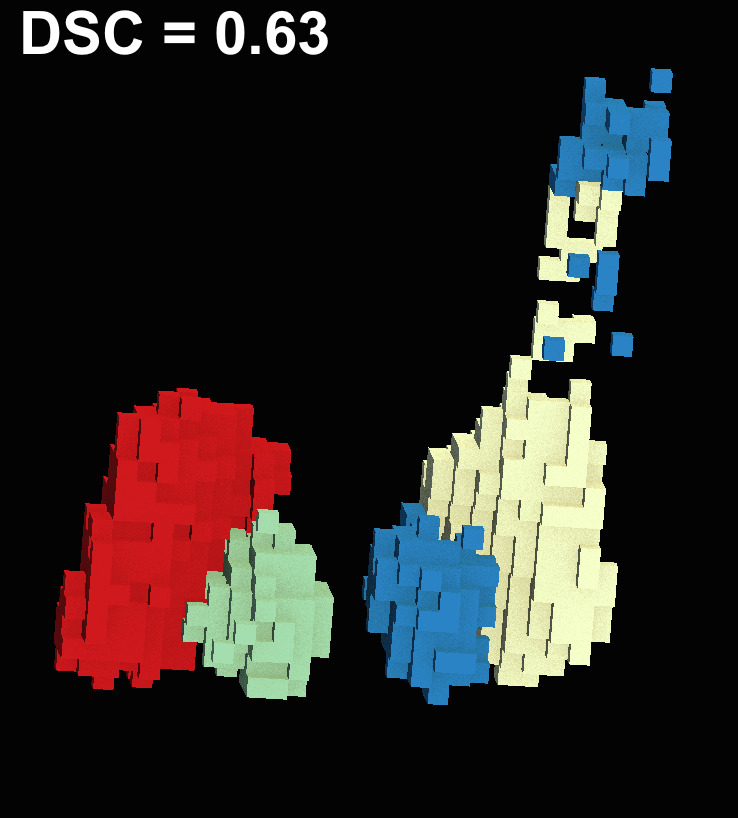} } \hfil
        \subfigure{\includegraphics[scale=\stnScale]{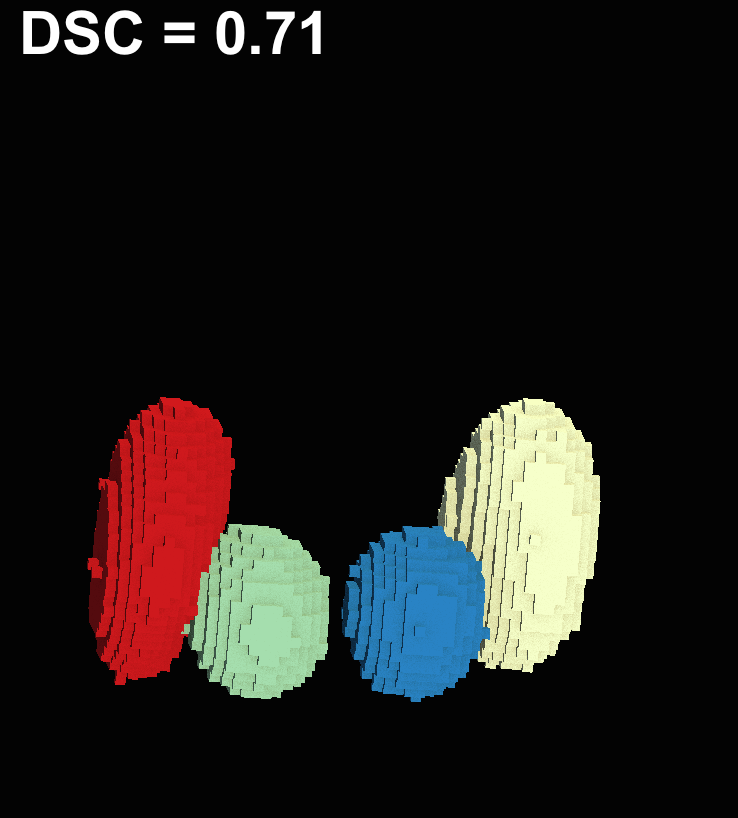}}%
     }%
     \vspace{-1cm}
     \centerline{ 
     \rotatebox{90}{$~~\qquad$worst}
        \subfigure{\includegraphics[scale=\stnScale]{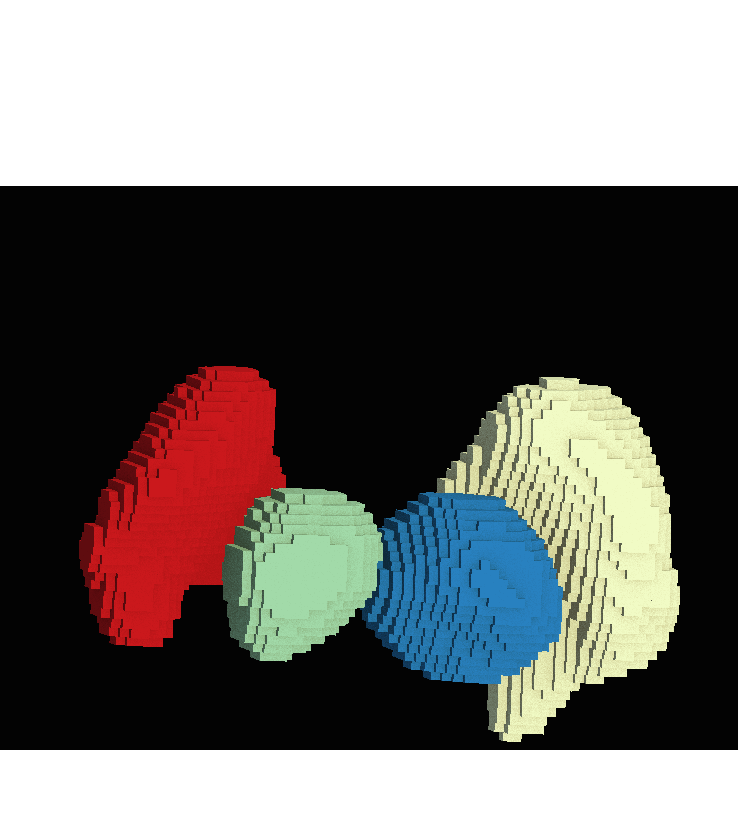} } \hfil
        \subfigure{\includegraphics[scale=\stnScale]{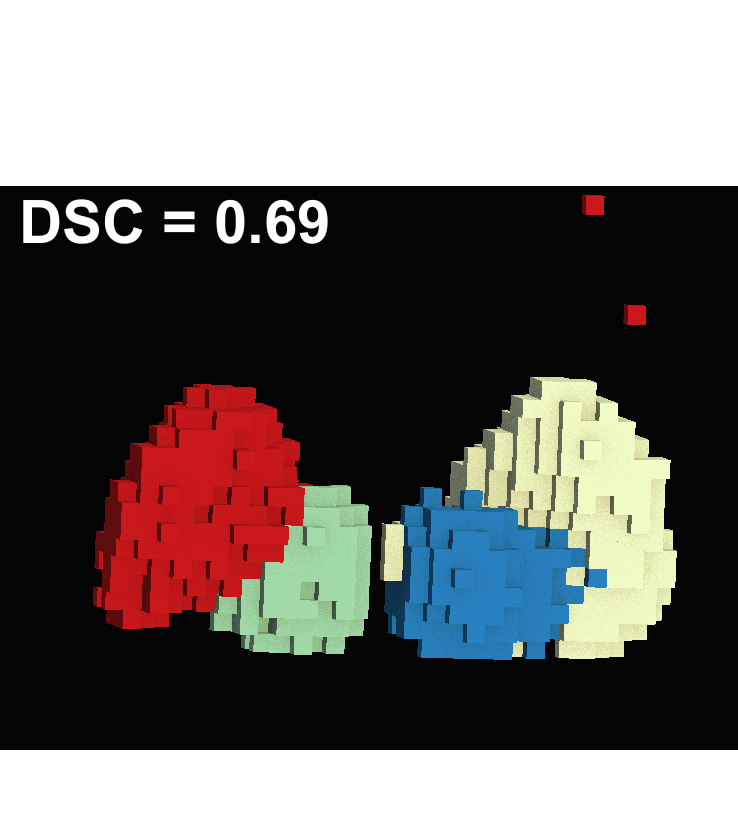} } \hfil
        \subfigure{\includegraphics[scale=\stnScale]{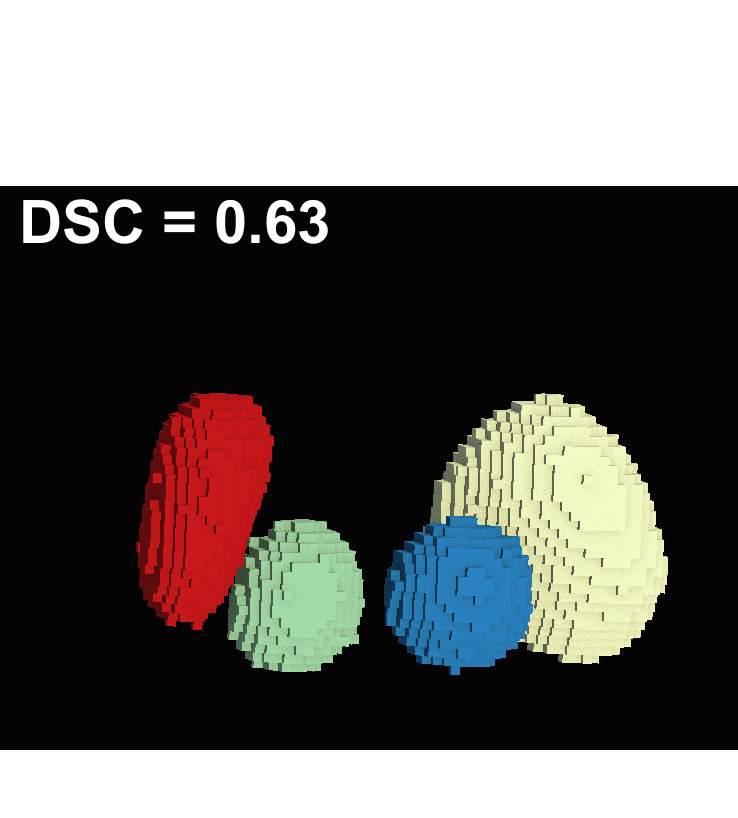}}%
     }%
    \vspace{-7mm}%
    \caption{Qualitative results for the brain structure segmentation experiments. Each row shows a different instance of results, where from top to bottom we present the \emph{best}, \emph{median} and \emph{worst} DSC differences (cf.~Fig.~\ref{stnDscDifferences}, right).}%
    \label{stnSeg}%
\end{figure} 
\vspace{-5mm}

\section{Conclusion}
We introduced the first combinatorial method for non-rigidly matching a 3D shape to a 3D image. The key idea is to represent the 3D shape as a triangular mesh and to solve a manifold-valued multi-labelling problem on the set of triangles. We determine an assignment of a rigid-body transformation associated with each triangle by minimising a cost function where the unary terms encode the local matching cost in the image and the pairwise terms penalise the amount of non-rigidity in the deformation. In particular, we propose an efficient discretisation of the unbounded 6-dimensional Lie group of rigid motions.  Moreover, we solve the large and NP-hard optimisation problem with a graph theoretic algorithm that is insensitive to initialisation and has the guarantee that the obtained solution is within a factor of the global optimum \cite{Boykov:2001ug}. 
Experimental validation confirms these benefits.

\subsection*{Acknowledgements}
The authors would like to thank \"Ozg\"un \c{C}i\c{c}ek for helpful feedback regarding the U-Net implementation, as well as Benedikt Staffler and Ankush Gupta for general feedback on CNNs.
CNN-related computations presented in this paper were carried out
using the HPC facilities of the University of Luxembourg~\cite{VBCG_HPCS14}.
The authors gratefully acknowledge the financial support by the Fonds National de la Recherche, Luxembourg (6538106, 8864515). This work was supported by the ERC Consolidator Grant 3D Reloaded.

{\small
\bibliographystyle{ieee}
\bibliography{references}
}

\end{document}